\documentclass[11pt]{article}

\usepackage[final]{acl}

 \usepackage{microtype}

\usepackage[english,bidi=default]{babel} 

\usepackage{hyperref}
\usepackage{url}
\usepackage{graphicx}
\usepackage{booktabs}
\usepackage{amsmath}
\usepackage{amssymb}
\usepackage{makecell}
\usepackage{subcaption}
\usepackage{todonotes}
\usepackage[table]{xcolor}
\usepackage{lipsum}

\title{What Matters When Linearizing Language Models?}
\title{What Matters in Linearizing Language Models? A Comparative Study of Architecture, Scale, and Task Adaptation}

\author{Patrick Haller \And Jonas Golde \\
\\
Humboldt-Universität zu Berlin
\\
\texttt{\{patrick.haller.1{\normalfont,} jonas.max.golde{\normalfont,} alan.akbik\}@hu-berlin.de} \And
Alan Akbik \\
}
\begin{document}

\maketitle
\begin{abstract}

Linearization has emerged as a strategy for developing efficient language models (LMs). Starting from an existing Transformer-based LM, linearization replaces the attention component with computationally efficient subquadratic \textit{token mixers}. However, as an increasing number of mixers are proposed, it remains unclear which inductive biases are best suited to inherit the original Transformer's capabilities. Furthermore, it is unknown how linearization is affected by parameter and token budget scaling. To address these questions, we propose a unified setup to compare seven representative architectures, including xLSTM, GLA, and Gated DeltaNet. Our findings reveal that performance hierarchies remain stable from 140M to 1.7B parameters, with error-correcting update rules demonstrating superior scaling exponents. We show that performance gaps are established early and persist through asymptotic maturity at 10B tokens, suggesting that state resolution is a more fundamental bottleneck than the distillation budget. Finally, while most models adapt to instruction tuning, only gated delta-rule formulations maintain the precision necessary for long-context retrieval, whereas additive models suffer from irreversible state saturation. These results suggest that for successful linearization, architectural inductive biases remain the primary constraint that cannot be overcome by simply scaling training compute.


\end{abstract}

\section{Introduction}

Recent research has increasingly focused on \textit{linearization} as a strategy for developing efficient language models~\citep{transformer_to_ssm,mercat2024linearizing,zhang2024lolcatslowranklinearizinglarge,wolf2020huggingfacestransformersstateoftheartnatural}. This process involves taking a pretrained Transformer and converting it into a subquadratic student model by replacing the original $O(L^2)$ self-attention layers with more efficient \textit{token mixers}, such as linear attention or gated recurrences. While the student typically retains the teacher's pretrained feed-forward weights, it must undergo a phase of \textit{knowledge distillation} to adapt the new token mixer to the teacher's learned representations. The goal of linearization 
is to produce models that maintain the performance of large-scale Transformers while offering greatly improved efficiency due to the linear-time inference and constant-memory overhead characteristics of subquadratic architectures.

However, as the variety of subquadratic token mixers has increased, it has become difficult to determine which specific inductive biases are most effective for successful linearization. Current literature often reports results under heterogeneous conditions—varying in model size, training budget, and distillation objectives—which makes it challenging to isolate whether observed gains stem from architectural design (such as the use of gating, decay, or the delta rule) or from the specific training recipe.

Beyond the choice of token mixer, critical questions remain regarding the scaling behavior and downstream viability of linearized models. It is currently unclear how the performance of these students scales with both parameter count and the number of distillation tokens, or whether the advantages of certain architectures persist as they grow from millions to billions of parameters. Furthermore, while the primary goal of linearization is to retain the capabilities of the teacher, the extent to which subquadratic students can successfully undergo instruction tuning or manage long-context retrieval after distillation remains an open area of inquiry.

\noindent
\textbf{A unified evaluation framework.} We address these gaps by conducting a systematic empirical study of seven representative subquadratic architectures, including xLSTM \citep{xlstm}, DeltaNet \citep{deltanet}, Gated DeltaNet \citep{yang2025gateddeltanetworksimproving}, GLA \citep{gla}, RetNet \citep{retnet}, and Kimi \citep{kimiteam2025kimilinearexpressiveefficient}. 

To isolate the token mixer as the primary variable, we develop a three-stage distillation pipeline that aligns attention-outputs and hidden-states while holding the teacher model, data distribution, and training recipe constant. By abstracting the distillation process from architecture-specific constraints, we provide each student with a dense supervision signal that enables a rigorous head-to-head comparison of their underlying state-update mechanisms.

Our study yields three primary insights into the linearization of language models:

\begin{itemize}
\item We scale student models from 140M to 1.7B parameters and observe that architectural performance hierarchies remain remarkably stable across scales. While gated architectures consistently outperform ungated baselines, we identify a crossover at the 1.7B scale where error-correcting update rules begin to demonstrate superior scaling exponents compared to purely additive alternatives.
    \item We evaluate the maturity of these architectures by extending distillation significantly beyond standard linearization recipes. Our findings indicate that relative performance gaps are established early in the training process and persist even as models approach their asymptotic limits, suggesting that the inherent state resolution of the token mixer is a more fundamental bottleneck than the distillation token budget.
    \item We evaluate the viability of linearized models for downstream and long-context tasks by performing instruction tuning on the distilled students. While we find that subquadratic models successfully adapt to instruction following, this process reveals a mixed picture regarding state resolution; specifically, most architectures suffer from a collapse in long-context retrieval, with only gated delta-rule formulations maintaining the precision required for long-range associative recall. 
   
\end{itemize}

Taken together, these findings establish a clear hierarchy for subquadratic design and demonstrate that architectural choices such as error-correcting update rules are more decisive for successful linearization than simply increasing the distillation token budget.

To facilitate further research into efficient architectures and ensure the reproducibility of our findings, we publicly release all base, instruction-tuned, and instruction-distilled models\footnote{\href{https://huggingface.co/collections/PatrickHaller/distillled-smollm2-17b}{Link} to HuggingFace Collection}.

\begin{table*}[h]
\centering
\footnotesize
\setlength{\tabcolsep}{4pt} 
\resizebox{\textwidth}{!}{
\begin{tabular}{lllll}
\toprule
\textsc{Architecture} & \textsc{Update Type} & \textsc{Input Dep.} & \textsc{State Update Rule} & \textsc{Granularity} \\
\midrule
\multicolumn{5}{c}{\textbf{Outer Product Based Matrix-Valued State} (Additive)} \\
\midrule
\textbf{Linear Attn} & Additive & - & $S_t = S_{t-1} + k_t v_t^\top$ & Ungated \\
\addlinespace
\textbf{RetNet} & Decay & No & $S_t = \gamma S_{t-1} + v_tk_t^\top$ & Channel-wise \\
\addlinespace
\textbf{GLA} & Gated Decay & Yes & $S_t = \mathrm{Diag}(\alpha_t) S_{t-1} + k_t v_t^\top$ & Channel-wise (diag.) \\
\addlinespace
\textbf{mLSTM (xLSTM)} & Gated Decay & Yes & $S_t = f_t S_{t-1} + i_t v_t k_t^\top$ & Head-wise Scalar \\
\midrule
\multicolumn{5}{c}{\textbf{Delta-Rule} (Overwriting)} \\
\midrule
\textbf{DeltaNet} & Error-Correction & Yes & $S_t = S_{t-1}(I - \beta_t k_t k_t^\top) + \beta_t v_t k_t^\top$ & Scalar Overwrite \\
\addlinespace
\textbf{Gated DeltaNet} & Error-Correction & Yes & $S_t = \alpha_t S_{t-1}(I - \beta_t k_t k_t^\top) + \beta_t v_t k_t^\top$ & Head-wise Scalar \\
\addlinespace
\textbf{Kimi (KDA)} & Error-Correction & Yes & $S_t = \mathrm{Diag}(\alpha_t) S_{t-1}(I - \beta_t k_t k_t^\top) + \beta_t v_t k_t^\top$ & Channel-wise (diag.) \\
\bottomrule
\end{tabular}}
\caption{
Taxonomy of subquadratic token mixers. We distinguish between additive accumulation, gated decay, and content-aware overwriting (Delta-rule). Formulas illustrate how the state $S_t \in \mathbb{R}^{d_k \times d_v}$ is updated per timestep.
}
\label{tab:gated_token_mixers}
\end{table*}

\section{Preliminaries and Related Work}

\subsection{Subquadratic Token Mixers}

Subquadratic token mixers replace quadratic self-attention with mechanisms that process sequences recursively by updating a compact state as tokens arrive. Rather than attending explicitly to all previous tokens, these models compress history into a learned state that is incrementally updated, enabling linear-time inference while shifting the burden of long-range dependency modeling onto the state update rule.

Across existing work, most subquadratic architectures differ along one central axis: how they retain, overwrite, or forget information over time. We group the models considered in this study according to this principle.

\noindent
\textbf{State accumulation via linear attention}. Linear attention approximates softmax attention by reformulating attention as a sum of key–value outer products stored in a state matrix. The state is updated additively at each step and queried by the current token, yielding linear complexity in sequence length but removing the explicit normalization and token-level selectivity of softmax attention.
A known limitation of this formulation is the lack of explicit forgetting, which can cause the state to become dominated by stale or irrelevant information. Several variants~\citep{rebased,hedgehog} address this by improving the feature map or kernel, but the underlying update remains purely additive.

\noindent
\textbf{Gated and decay-based state updates.} An alternative design introduces explicit control over memory retention through decay or gating. Rather than accumulating all past information, these models modulate the previous state before adding new content. RetNet~\citep{retnet} applies a uniform decay to the entire state, biasing the model toward recent context, Gated Linear Attention (GLA)~\citep{gla} refines this idea by applying channel-wise decay, allowing different components of the state to be forgotten at different rates.
More expressive variants introduce input and forget gates, drawing inspiration from recurrent neural networks. Models such as xLSTM~\citep{xlstm} update their state using multiplicative gates that regulate both how much new information is written and how much of the previous state is preserved. This enables selective retention and controlled overwriting.
\noindent
\textbf{Delta-rule and content-based overwriting.}
Delta-rule models~\citep{deltanet} adopt a content-aware update mechanism that partially overwrites state components aligned with the current input. This induces a form of last-write-wins memory, where new evidence replaces older, related information. Gated variants, such as Gated DeltaNet~\citep{yang2025gateddeltanetworksimproving} and Kimi~\citep{kimiteam2025kimilinearexpressiveefficient}, combine this overwrite rule with decay terms, providing fine-grained control over memory retention. Rather than approximating attention kernels, these models replace token-level selection with structured state updates.

\subsection{Linearizing Pretrained Transformers}

Rather than training subquadratic language models from scratch, a growing body of work studies how to \emph{linearize} pretrained Transformers by replacing softmax attention with more efficient token mixers and transferring knowledge via distillation~\citep{hinton2015distillingknowledgeneuralnetwork}. Early approaches focused on fine-tuning pretrained models with recurrent or decaying attention mechanisms~\citep{kasai2021finetuningpretrainedtransformersrnns,mao2022finetuningpretrainedtransformersdecaying}, demonstrating that pretrained representations can be partially preserved despite architectural changes.

More recent work has proposed increasingly structured linearization pipelines. SUPRA~\citep{Mercat2024LinearizingLL} introduces a scalable uptraining framework that converts pretrained Transformers into recurrent architectures through progressive adaptation. LoLCATs~\citep{zhang2024lolcatslowranklinearizinglarge} combines low-rank adaptation~\citep{hu2021loralowrankadaptationlarge} with attention transfer to approximate softmax attention efficiently. MOHAWK~\citep{transformer_to_ssm} proposes a staged distillation procedure that aligns materialized attention maps between teacher and student, enabling the transfer of quadratic attention behavior into subquadratic models.


Subsequent extensions integrate architectural reuse and instruction tuning. Mamba-LLaMA~\citep{wang2025mamballamadistillingaccelerating} applies progressive distillation to Mamba-based students, LIGER~\citep{lan2025ligerlinearizinglargelanguage} constructs gating modules with additional mechanisms such as sliding-window attention, and \citet{yueyu2025arwkvpretrainneedrnnattentionbased} linearize Qwen-2.5 using RWKV-style blocks with hidden-state alignment. LIZARA~\citep{vannguyen2025lizardefficientlinearizationframework} further augments subquadratic token mixers with adaptive memory modules.


In contrast to prior work that typically targets a single student architecture or emphasizes attention-map reconstruction, our study adopts an \emph{architecture-agnostic} linearization framework based on attention-output and hidden-state alignment. This enables direct comparison of diverse subquadratic token mixers under identical training conditions, isolating the effects of state-update mechanisms, gating, and scaling behavior. As Mamba-style~\citep{mamba2} models have already been extensively explored, we focus on alternative architectures spanning additive, gated, and delta-rule–based update families.

\section{Linearization Framework}

We train subquadratic student models via a three-stage distillation pipeline that progressively aligns the student with a pretrained Transformer teacher. The pipeline is designed to stabilize optimization, isolate the token mixer as the primary source of variation, and enable fair comparison across architectures. All student models are initialized by copying all parameters from the teacher model, while newly introduced parameters are initialized from scratch.

In contrast to prior approaches such as MOHAWK, which directly align the teacher’s attention maps with materialized attention maps of the student, we align the full attention output produced by the token mixer. Direct attention-map alignment is inherently tied to attention-specific inductive biases and primarily constrains query–key interactions. Aligning the complete attention output instead provides a denser supervision signal and is model-agnostic, as it remains well-defined across linear attention, gated recurrent, and state-update–based token mixers.

\paragraph{Stage 1: Token-mixer (attention-output) alignment.}
In the first stage, we align the student’s token mixer with the teacher’s attention output.
For each aligned layer $\ell$, we minimize a mean squared error loss
\begin{equation}
\mathcal{L}_{\mathrm{AO}}^{\ell}
=
\left\| a_S^{\ell}(x) - a_T^{\ell}(x) \right\|_2^2 .
\end{equation}
The overall attention-output alignment objective is
\begin{equation}
\mathcal{L}_{\mathrm{AO}}
=
\frac{1}{|\mathcal{L}_1|}
\sum_{\ell \in \mathcal{L}_1}
\mathcal{L}_{\mathrm{AO}}^{\ell},
\end{equation}
where $\mathcal{L}_1$ denotes the set of aligned layers.
This stage directly supervises the complete token-mixing pathway, including value projections,
output projections, and gating mechanisms.

\paragraph{Stage 2: Hidden-state alignment.}
In the second stage, we align intermediate hidden representations of teacher and student.
For each selected layer $\ell$, we minimize the mean squared $\ell_2$ distance
\begin{equation}
\mathcal{L}_{\mathrm{H2H}}^{\ell}
=
\frac{1}{T}
\sum_{t=1}^{T}
\left\|
h_S^{\ell}(x)_t - h_T^{\ell}(x)_t
\right\|_2 .
\end{equation}
Aggregating over a set of layers $\mathcal{L}_2$, the hidden-state alignment loss is
\begin{equation}
\mathcal{L}_{\mathrm{H2H}}
=
\frac{1}{|\mathcal{L}_2|}
\sum_{\ell \in \mathcal{L}_2}
\mathcal{L}_{\mathrm{H2H}}^{\ell}.
\end{equation}

\paragraph{Stage 3: End-to-end knowledge distillation.}
In the final stage, we train the student end-to-end using standard knowledge distillation.
The objective combines next-token prediction with distribution matching:
\begin{equation}
\mathcal{L}_{\mathrm{KD}}
=
\mathcal{L}_{\mathrm{CE}}
+
\lambda_{\mathrm{KL}} \mathcal{L}_{\mathrm{KL}},
\end{equation}

where the distillation loss is defined as
\begin{equation}
\mathcal{L}_{\mathrm{KL}}
=
\frac{1}{T}
\sum_{t=1}^{T}
\mathrm{KL}
\big(
p_T(\cdot \mid x_{\le t})
\;\|\;
p_S(\cdot \mid x_{\le t})
\big).
\end{equation}

The three objectives are applied sequentially. This staged optimization improves stability and prevents interference between alignment signals, enabling controlled comparison of subquadratic token mixers under a unified training protocol.

\section{Experiments}

We select seven representative subquadratic token mixers to enable controlled comparisons along two key architectural axes: the update rule and the presence and granularity of gating. Specifically, we consider two dominant update families - \textit{outer-product additive updates} and \textit{delta-rule–based updates} - and, within each family, architectures with head-wise scalar gating as well as channel-wise gating.

Within the outer-product family, we include Linear Attention as an ungated baseline, GLA~\citep{gla} as a channel-wise gated variant, and xLSTM~\citep{xlstm} as a head-wise scalar gated model with input-dependent gating. Within the delta-rule family, we include DeltaNet~\citep{deltanet} as a channel-wise gated model, Gated DeltaNet~\citep{yang2025gateddeltanetworksimproving} as a head-wise scalar gated variant and Kimi~\citep{kimiteam2025kimilinearexpressiveefficient}
 as a channel-wise gated variant. In addition, we include RetNet~\citep{retnet}, which employs a head-wise scalar but input-independent retention gate and thus occupies a distinct position between ungated and input-dependent gated models.

This selection spans ungated, input-independent gated, and input-dependent gated architectures across both update families, enabling systematic analysis of which inductive biases most strongly affect distillation and instruction tuning performance. We validate the computational efficiency of all subquadratic models in Appendix~\ref{appendix:bench}.

\subsection{Experiment 1: Comparison Across Model Size} Our first inquiry establishes the relative ranking of these eight subquadratic architectures under a unified framework and investigates whether these performance hierarchies remain stable as model capacity increases. 


\subsubsection{Experimental Setup} 
\textbf{Token budget.} We train all architectures on a fixed budget of 3B tokens sampled from \textsc{FineWeb}~\citep{fineweb}. Prior work has shown that a 3B token budget yields state-of-the-art models through linearization~\citep{transformer_to_ssm}. Text is concatenated and chunked into fixed-length sequences of 512 tokens. Following the distillation protocol of~\citet{transformer_to_ssm}, we allocate fixed token budgets for two alignment objectives: 80M tokens for matrix-mixing distillation and 160M for hidden-state alignment.

\begin{table}[t]
\centering
\small
\setlength{\tabcolsep}{8pt} 
\resizebox{\linewidth}{!}{
\begin{tabular}{l ccc}
    \toprule
    \textsc{Model} & \textsc{Avg.} & \textsc{Recov.} & \textsc{Rank} \\
    \midrule
    \rowcolor[gray]{.97} \multicolumn{4}{l}{\textbf{Scale: $\sim$140M Parameters}} \\
    \textit{Teacher: SmolLM2-135M} & \textit{44.6} & \textit{100\%} & -- \\
    xLSTM & 41.3 & 92.6\% & 1 \\
    Kimi & 40.3 & 90.2\% & 2 \\
    Gated DeltaNet & 39.5 & 88.5\% & 3 \\
    GLA & 39.1 & 87.6\% & 4 \\
    RetNet & 38.4 & 86.1\% & 5 \\
    DeltaNet & 38.0 & 85.1\% & 6 \\
    Linear Attn & 32.7 & 73.2\% & 7 \\
    \midrule
    \rowcolor[gray]{.97} \multicolumn{4}{l}{\textbf{Scale: $\sim$360M Parameters}} \\
    \textit{Teacher: SmolLM2-360M} & \textit{53.1} & \textit{100\%} & -- \\
    xLSTM & 47.7 & 89.8\% & 1 \\
    Gated DeltaNet & 46.3 & 87.2\% & 2 \\
    GLA & 45.6 & 85.8\% & 3 \\
    Kimi & 43.9 & 82.7\% & 4 \\
    RetNet & 43.7 & 82.2\% & 5 \\
    DeltaNet & 41.5 & 78.0\% & 6 \\
    Linear Attn & 37.2 & 70.1\% & 7 \\
    \midrule
    \rowcolor[gray]{.97} \multicolumn{4}{l}{\textbf{Scale: $\sim$1.7B Parameters}} \\
    \textit{Teacher: SmolLM2-1.7B} & \textit{62.9} & \textit{100\%} & -- \\
    Gated DeltaNet & 59.2 & 94.2\% & 1 \\
    xLSTM & 57.6 & 91.6\% & 2 \\
    DeltaNet & 55.7 & 88.6\% & 3 \\
    RetNet & 54.8 & 87.1\% & 4 \\
    GLA & 54.4 & 86.5\% & 5 \\
    Kimi & 51.5 & 82.0\% & 6 \\
    Linear Attn & 44.6 & 70.9\% & 7 \\
    \bottomrule
\end{tabular}}
\caption{Summary of zero-shot downstream performance across scales. Recovery (Recov.) denotes the percentage of teacher performance retained by the linearized student. Models are ranked per scale by average performance.}
\label{tab:summary_scaling}
\end{table}

\begin{figure}
    \centering
    \includegraphics[width=\linewidth]{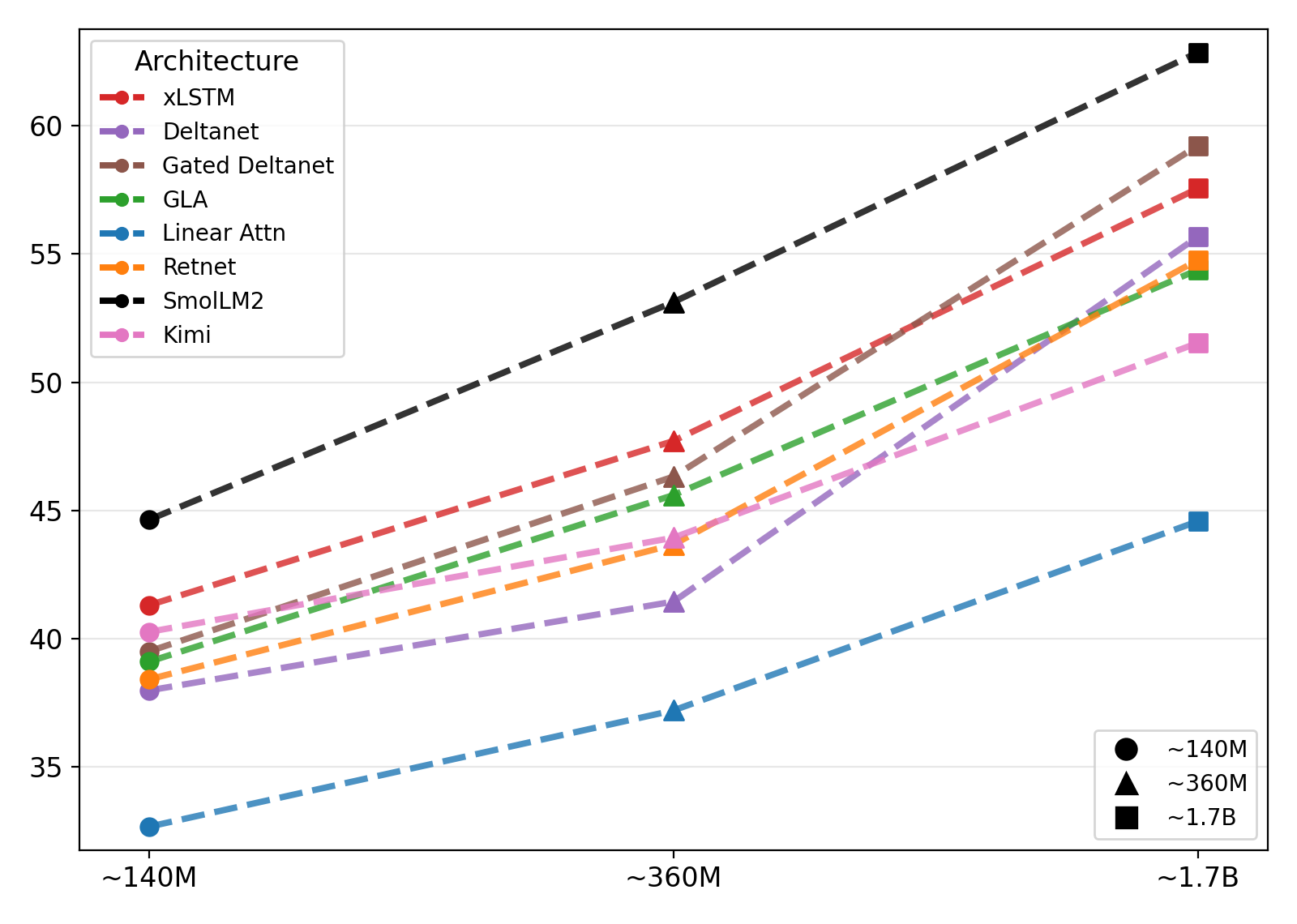}
    \caption{Average evaluation score for all 7 (+ teacher) considered token mixers at three model sizes. All architectures markedly improve at larger model sizes, though some architectures, such as Gated DeltaNet, benefit most from model scaling.}
    \label{fig:param_scaling}
\end{figure}

\begin{figure*}[t]
    \centering
    \includegraphics[width=\linewidth]{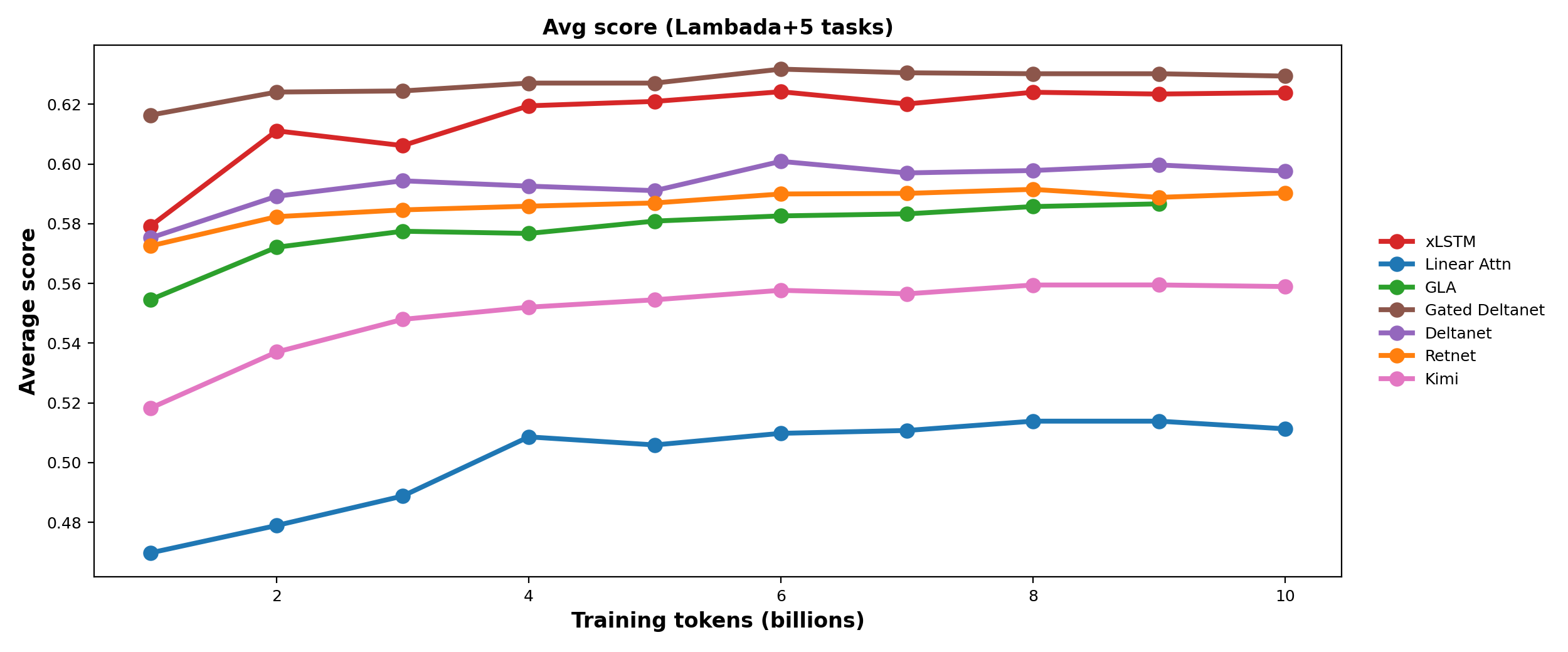}
    \caption{Token scaling behavior during training. We evaluate each 1B tokens of training and report average scores across all benchmarks.}
    \label{fig:token_scaling}
\end{figure*}
\noindent 
\textbf{Model sizes.} We instantiate all seven subquadratic architectures at three scales: 140M, 360M, and 1.7B parameters. As teachers, we use the SmolLM2 family~\citep{allal2025smollm2smolgoesbig}, a series of Llama-style Transformer models. Each teacher corresponds in parameter count to our student sizes: SmolLM2-135M, SmolLM2-360M, and SmolLM2-1.7B\footnote{Note that since specific token mixers utilize different internal layer structures, parameter counts are matched as closely as possible but are not exactly identical across architectures.}.

\noindent\textbf{Evaluation.} Student and teacher models are evaluated on seven zero-shot tasks using the LM-Eval-Harness~\citep{eval-harness} and LM-Pub-Quiz~\citep{ploner2024lmpubquizcomprehensiveframeworkzeroshot} frameworks: LAMBADA~\citep{lambada}, WinoGrande~\citep{winogrande}, ARC-Easy and ARC-Challenge~\citep{arc_easy}, PIQA~\citep{piqa}, HellaSwag~\citep{zellers-etal-2019-hellaswag} and BEAR~\citep{wiland2024bearunifiedframeworkevaluating}. LAMBADA results are reported as the mean of its Standard and OpenAI variants.

\subsubsection{Results} 
Our results are summarized in Figure~\ref{fig:param_scaling} and Table~\ref{tab:summary_scaling}. As Figure~\ref{fig:param_scaling} shows, we observe smooth, monotonic scaling across all eight architectures with no signs of divergence or early saturation. This confirms that the considered subquadratic models can be trained stably via distillation across a wide parameter range.

However, as Table~\ref{tab:summary_scaling} shows, 
scaling efficiency—the ability to close the performance gap to the Transformer teacher—differs markedly between architectures. Gated state-update models, such as xLSTM and Gated DeltaNet, consistently define the Pareto frontier across all scales. A key discovery in our comparison is the shift in the performance hierarchy at the largest scale: while xLSTM maintains the highest recovery at 140M and 360M, we observe a notable "crossover" at 1.7B, where Gated DeltaNet emerges as the top performer with 94.2\% recovery. This suggests that error-correction mechanisms, such as the Delta-rule, may possess superior scaling laws as model capacity increases.

Among attention-inspired approaches, GLA and RetNet exhibit nearly identical scaling trajectories, with GLA maintaining a consistent but narrow lead. Conversely, feature-map-based Linear Attention lags substantially behind. Critically, the performance gap between Linear Attention and gated models widens as size increases (from a $\sim$14\% gap at 140M to $\sim$20\% at 1.7B), suggesting that its deficiencies are intrinsic to the architecture and cannot be overcome simply by increasing parameter count.

\subsection{Experiment 2: Asymptotic Performance and State Resolution}

While Experiment 1 established scaling laws relative to parameter count, it did not address whether weaker architectures simply require more data to converge. In this experiment, we evaluate the "maturity" of these models by fixing the size at $\sim$1.7B parameters and extending training to 10B tokens. This allows us to observe both the asymptotic knowledge limit (via downstream benchmarks) and the state resolution (via retrieval tasks) of each architecture.

\subsubsection{Asymptotic Knowledge Scaling}

We train over 10B tokens sampled from \textsc{FineWeb} and use the same token splits for the three linearization phases. We checkpoint the models after each 1B tokens of training. Each checkpoint is evaluated on the same benchmarks as in Experiment 1. 

\noindent 
\textbf{Results.} As shown in Figure~\ref{fig:token_scaling}, all architectures benefit from additional training. However, the results confirm that more data does not close the architectural gap.

Gated state-update models, specifically Gated DeltaNet and xLSTM, maintain their lead throughout training, with performance largely plateauing after 6--7B tokens. 
In contrast, ungated models such as RetNet and DeltaNet saturate earlier and at a significantly lower level. This suggests that the performance differences observed in Experiment 1 are not a consequence of undertraining, but rather represent fundamental limits in how different token mixers compress information into their recurrent states.

\begin{table*}[ht]
    \centering
    \small
    \begin{tabular}{l cccc @{\hspace{15pt}} cccc @{\hspace{15pt}} cccc}
        \toprule
        & \multicolumn{4}{c}{\textsc{S-NIAH-1} (Pass-key)} & \multicolumn{4}{c}{\textsc{S-NIAH-2} (Number)} & \multicolumn{4}{c}{\textsc{S-NIAH-3} (UUID)} \\
        \cmidrule(r{15pt}){2-5} \cmidrule(r{15pt}){6-9} \cmidrule{10-13}
        Model & 512 & 1k & 2k & 4k & 512 & 1k & 2k & 4k & 512 & 1k & 2k & 4k \\
        \midrule
        Gated DeltaNet & \textbf{99.4} & \textbf{97.8} & \textbf{81.8} & \textbf{29.8} & \textbf{99.0} & \textbf{93.4} & \textbf{55.0} & \textbf{15.6} & \textbf{16.4} & \textbf{3.2} & \textbf{1.2} & \textbf{0.4} \\
        xLSTM & 95.8 & 78.0 & 25.2 & 8.4 & 98.6 & 82.2 & 32.0 & 8.6 & 6.2 & 0.4 & 0.0 & 0.0 \\
        DeltaNet & 98.8 & 96.2 & 0.0 & 0.0 & 99.2 & 77.0 & 24.6 & 0.0 & 1.8 & 0.0 & 0.0 & 0.0 \\
        Kimi & 33.4 & 7.4 & 2.0 & 0.8 & 91.8 & 15.2 & 7.8 & 3.4 & 0.0 & 0.0 & 0.0 & 0.0 \\
        GLA & 10.6 & 1.0 & 0.0 & 0.0 & 50.0 & 13.0 & 3.0 & 1.6 & 0.0 & 0.0 & 0.0 & 0.0 \\
        RetNet & 10.4 & 0.0 & 0.0 & 0.0 & 30.2 & 10.2 & 0.0 & 0.0 & 0.0 & 0.0 & 0.0 & 0.0 \\
        Linear Attn & 0.0 & 0.0 & 0.0 & 0.0 & 0.0 & 0.0 & 0.0 & 0.0 & 0.0 & 0.0 & 0.0 & 0.0 \\
        \bottomrule
    \end{tabular}
    \caption{Needle-In-A-Haystack (NIAH) performance of distilled models at 10B tokens. Scores represent retrieval accuracy (\%). Gated DeltaNet demonstrates significantly higher state resolution at extended context lengths.}
    \label{tab:niah_distilled}
\end{table*}

\subsubsection{Retrieval and State Resolution}

A model's ability to accumulate knowledge over 10B tokens is inextricably linked to its ability to manage its internal state. To probe the "resolution" of these states, we evaluate the 10B-token checkpoints on three "Needle-In-A-Haystack" (NIAH) tasks~\citep{hsieh2024rulerwhatsrealcontext}. In NIAH, models are tasked to retrieve a specific piece of information (the "needle") from a long sequence of tokens. 

\noindent \textbf{Results.} The retrieval results in Table~\ref{tab:niah_distilled} reveal a pervasive collapse in state resolution across most subquadratic architectures as sequence length increases. While some models achieve high accuracy at a 512-token context, performance degrades rapidly at medium sequence lengths. Linear Attention and RetNet fail almost entirely beyond the 512-token mark, while competitive gated models like xLSTM and DeltaNet experience a dramatic loss in precision at 2k tokens, dropping to 25.2\% and 0.0\% respectively on S-NIAH-1. This suggests that for associative or simple gated mixers, the fixed-size state becomes "saturated" or "noisy" as it accumulates more information, lacking the necessary precision for long-range associative recall.

In contrast, Gated DeltaNet exhibits significantly more robust state-management compared to the other subquadratic alternatives. It is the only architecture in our study to preserve a degree of retrieval success at 2k and 4k contexts, achieving 81.8\% accuracy on S-NIAH-1 at 2k tokens—a substantial margin over the next-best student model (xLSTM). While this still falls short of the near-perfect retrieval typically observed in Transformer teachers with explicit KV-caches, the divergence indicates that the gated Delta-rule provides a more effective mechanism for targeted state updates. 

\subsection{Experiment 3: Instruction Adaption: Post-Training vs. Instruction-Aware Distillation}

\begin{table*}[ht]
\centering
\small
\begin{tabular*}{\textwidth}{@{\extracolsep{\fill}} l c | c | cccc }
    \toprule
    & \textsc{Zero-Shot} & \textsc{Retrieval} & \multicolumn{4}{c}{\textsc{Instruction Following \& Reasoning}} \\
    \cmidrule{4-7}
    \textsc{Model} & \makecell{\textsc{LLM Avg}} & \makecell{\textsc{NIAH-S Avg}} & \textsc{IFEval} & \textsc{GSM8K} & \textsc{SWDE} & \textsc{LongB.} \\
    \midrule
    
    \multicolumn{7}{l}{\textbf{Regime A: Post-Training Adaptation} (Teacher: \textit{SmolLM2-1.7B} + \textit{SmolTalk} SFT 1 Epoch)} \\
    \midrule
    \textit{Teacher Baseline} & \textit{70.24} & \textit{99.31} & \textit{56.71} & \textit{42.08} & \textit{23.85} & \textit{17.45} \\
    \midrule
Gated DeltaNet & \textbf{61.07} & \textbf{57.33} & 40.89 & 8.28 & 13.33 & \textbf{17.74} \\
    xLSTM          & 60.52 & 47.15 & \textbf{45.08} & \textbf{10.17} & 14.03 & 16.68 \\
    GLA            & 56.59 & 10.78 & 40.89 & 8.28 & 13.33 & \textbf{17.74} \\
    DeltaNet       & 56.29 & 39.38 & 36.21 & 9.72 & 3.62 & 2.27 \\
    RetNet         & 56.29 & 6.28  & 32.61 & 6.39 & 9.67 & 2.05 \\
    Kimi           & 50.68 & 17.36 & 32.61 & 8.01 & \textbf{14.31} & 6.82 \\
    Linear Attn    & 44.70 & 0.00  & 24.46 & 1.62 & 2.21 & 0.00 \\
    
    \midrule \midrule
    
    \multicolumn{7}{l}{\textbf{Regime B: Instruction-Aware Distillation} (Teacher: \textit{SmolLM2-1.7B-Instruct})} \\
    \midrule
    \textit{Teacher Baseline} & \textit{70.57} & \textit{98.06} & \textit{60.19} & \textit{49.51} & \textit{22.32} & \textit{25.64} \\
    \midrule
    Gated DeltaNet & \textbf{55.35} & \textbf{47.33} & \textbf{47.12} & \textbf{20.02} & \textbf{10.08} & \textbf{15.24} \\
    Deltanet       & 55.34 & 40.10 & 42.17 & 13.12 & 9.36 & 13.23 \\
    RetNet         & 53.69 & 10.61 & 31.06 & 2.81 & 7.11 & 7.98 \\
    xLSTM          & 51.64 & 10.41 & 41.73 & 13.50 & 8.10 & 14.91 \\
    GLA            & 49.11 & 4.35  & 29.50 & 5.84 & 5.13 & 10.79 \\
    Kimi           & 44.38 & 4.48 & 30.58 & 6.07 & 6.30 & 8.91 \\
    Linear Attn    & 31.28 & 0.00 & 22.78 & 0.00 & 1.40 & 2.00 \\
    \bottomrule
\end{tabular*}
\caption{Comparison of instruction-following performance across two adaptation strategies. Regime A evaluates the effectiveness of standard post-training SFT on linearized base models, while Regime B examines a holistic instruction-aware distillation pipeline. Each regime is contrasted against its respective Transformer teacher to measure relative recovery of specialized capabilities.}
\label{tab:instruction_results_wide}
\end{table*}

A key open question in linearizing pretrained language models is whether subquadratic students can successfully acquire instruction-following behaviour, and how this capability depends on the timing and role of distillation. Prior work has explored instruction tuning either as a post-training adaptation step or as part of the distillation pipeline from instruction-tuned teachers. We comparatively evaluate both setups
to isolate the effect of instruction supervision versus instruction-aware distillation

\subsubsection{Experimental Setup}

We evaluate instruction adaption at the \textasciitilde1.7B parameter scale, starting from student models trained for 10B tokens as described in Experiment 2. We consider two complementary adaption regimes:

\begin{enumerate}
    \item \textbf{Post-training instruction tuning:} Linearized models are fine-tuned directly on the \textit{SmolTalk} dataset, containing 1.1M instruction-response pairs (~1B tokens), using the same optimization recipe~\footnote{Recipe:~\url{https://github.com/huggingface/alignment-handbook/tree/main/recipes/smollm2}} employed to train \textit{SmolLM2-1.7B-Instruct}~\citep{allal2025smollm2smolgoesbig}. No teacher model or distillation loss is used in this setting.
    \item  \textbf{Instruction-aware distillation:}
Students are initialized from \textit{SmolLM2-1.7B-Instruct} and undergo Stage 1 and Stage 2 distillation (1B tokens total) followed by instruction tuning with a Stage 3 distillation loss on \textit{SmolTalk}. 
\end{enumerate}
\noindent 

All adapted models are evaluated on both zero-shot downstream benchmarks and instruction-following tasks, including IFEval~\citep{zhou2023instructionfollowingevaluationlargelanguage}, GSM8K~\citep{cobbe2021training}, SWDE and LongBench~\citep{bai2024longbench}\footnote{To evaluate long-context capabilities, we include five subsets from LongBench~\citep{bai2024longbench2}: WikiMQA, MultiFieldQA, NarrativeQA, TREC, and TriviaQA.}, enabling direct comparison between adaption strategies.

\subsubsection{Results}

\noindent \textbf{Post-training instruction tuning amplifies architectural differences.}
Instruction tuning consistently improves performance across all subquadratic architectures, demonstrating that linearized models remain trainable after distillation. However, the magnitude of improvement varies systematically with the underlying token mixer. Architectures that already exhibit strong pretraining performance, most notably gated state-update models such as xLSTM and Gated DeltaNet, benefit substantially more from instruction tuning, while weaker architectures show limited gains. As a result, instruction tuning amplifies existing performance differences rather than reducing them

\noindent \textbf{Instruction-aware distillation is inconsistent and degrades long-context performance.}
Instruction-aware distillation from an instruction-tuned teacher does not provide a consistent advantage over the pretraining-plus-post-training setup. While chat distillation can outperform direct instruction tuning on selected downstream instruction benchmarks, its effects vary across architectures. More critically, instruction-distilled models exhibit substantially worse performance on long-context retrieval benchmarks such as NIAH and LongBench. These degradations mirror the long-context failures observed prior to instruction adaptation, indicating that instruction-aware distillation does not compensate for insufficient state resolution in the student architecture.
\noindent
Taken together, these results show that instruction adaptation does not alter the fundamental architectural constraints imposed during linearization. Post-training instruction tuning reliably improves performance but primarily benefits already strong models, while instruction-aware distillation introduces additional variance without improving long-context robustness. The ability to selectively retain and overwrite information remains the dominant factor governing instruction-following performance in linearized language models.

\section{Conclusion}
%
%
%
We present a systematic empirical evaluation of subquadratic language models trained via knowledge distillation from Transformer teachers. By comparing diverse token mixers under a unified training and evaluation setup, we isolate the effect of architectural design from confounding factors such as data, optimization, and post-training procedures.

Our results show that while subquadratic models can be trained stably across scales, architectural choices induce persistent performance gaps. Models based on delta-rule updates and gated recurrent mechanisms consistently outperform outer-product–based linear attention. In particular, head-wise or scalar gating enables more effective memory control and long-range information retention than channel-wise diagonal gating.

We further find that post-training adaptation, including instruction tuning and instruction-level distillation, amplifies pretraining differences rather than compensating for weaker inductive biases. This suggests that selective memory and controlled forgetting are prerequisites for effective downstream alignment, rather than properties recoverable through supervision alone.

Overall, our findings indicate that the success of linearizing language models depends primarily on the structure of the token mixer, providing a clearer empirical basis for designing efficient alternatives to quadratic attention. By releasing our models and findings, we aim to support further progress in subquadratic language modeling.

\section*{Limitations}
While our study provides a comprehensive empirical comparison of subquadratic language models, several limitations should be noted.
\\
First, we restrict our analysis to a \textbf{fixed Transformer teacher family}. Although this enables controlled comparisons across student architectures, different teacher models or pretraining objectives may induce different transfer dynamics.
\\\\
Second, we do not investigate \textbf{hybrid architectures} (intra- or inter-layer) that combine multiple token mixers within or across layers.
Such designs would substantially expand the design space beyond the scope of this work. Moreover, while hybrid token-mixer architectures are an active area of research, prior work suggests that hybrid performance cannot be predicted from standalone components \citep{wang2025systematicanalysishybridlinear}, motivating our focus on isolating single-mixer architectural effects.
\\\\
Finally, our analysis focuses primarily on downstream performance and scaling behavior. We do not conduct a detailed \textbf{mechanistic or qualitative analysis} of internal dynamics, such as gating behavior, which could provide additional insight into why certain architectures align more effectively than others.

\bibliography{main}

\appendix

\section{Full Results: Experiment 1}\label{appendix:parameter_scaling}

This section reports full downstream evaluation results for the model size scaling experiment (Experiment 1). All models were trained on 3B distillation tokens. We include per-model and per-task scores for all student architectures at each parameter scale, corresponding to the averaged results summarized in the main text.

\begin{table*}[h]
\centering
\setlength{\tabcolsep}{4.5pt}
\resizebox{\textwidth}{!}{
\begin{tabular}{l ccccccc rr}
    \toprule
    \textsc{Model} & \makecell{\textsc{Lamb.} \\ \tiny acc.} & \makecell{\textsc{Wino.} \\ \tiny acc.} & \makecell{\textsc{Arc-E} \\ \tiny acc.n.} & \makecell{\textsc{Arc-C} \\ \tiny acc.n.} & \makecell{\textsc{Piqa} \\ \tiny acc.n.} & \makecell{\textsc{Hell.} \\ \tiny acc.n.} & \makecell{\textsc{BEAR} \\ \tiny acc.} &  \textsc{Avg.} & \textsc{Recov.} \\
    \midrule
    \rowcolor[gray]{.97} \multicolumn{9}{l}{\textbf{Scale: $\sim$140M Parameters}} \\
    SmolLM2-135M (Teacher) & 39.16 & 51.93 & 58.67 & 29.61 & 68.01 & 43.23 & 21.84 & 44.64 & 100.0\% \\
    Gated DeltaNet & 22.92 & 51.85 & 51.89 & 28.24 & 65.40 & 38.10 & 18.17 & 39.51 & 88.52\% \\
    xLSTM & 27.39 & 50.91 & 54.12 & 28.84 & 66.54 & 39.10 & 22.28 & 41.31 & 92.55\% \\
    Kimi & 22.28 & 53.20 & 51.47 & 28.58 & 65.45 & 37.81 & 23.11 & 40.27 & 90.22\% \\
    GLA & 22.04 & 52.25 & 51.52 & 27.13 & 65.29 & 37.19 & 18.35 & 39.11 & 87.62\% \\
    RetNet & 17.02 & 53.35 & 49.96 & 26.54 & 64.69 & 35.81 & 21.54 & 38.42 & 86.06\% \\
    DeltaNet & 17.09 & 51.38 & 49.96 & 27.90 & 65.29 & 35.66 & 18.63 & 37.99 & 85.10\% \\
    Linear Attn & 9.51 & 50.75 & 42.85 & 22.35 & 61.86 & 30.88 & 10.56 & 32.68 & 73.21\% \\
    \midrule
    \rowcolor[gray]{.97} \multicolumn{9}{l}{\textbf{Scale: $\sim$360M Parameters}} \\
    SmolLM2-360M (Teacher) & 49.29 & 58.72 & 68.22 & 37.88 & 71.98 & 56.33 & 29.45 & 53.12 & 100.0\% \\
    Gated DeltaNet & 31.79 & 55.17 & 60.19 & 32.34 & 69.80 & 49.80 & 25.14 & 46.32 & 87.19\% \\
    xLSTM & 37.30 & 56.43 & 60.06 & 32.76 & 71.38 & 51.24 & 24.80 & 47.71 & 89.81\% \\
    Kimi & 26.96 & 52.01 & 57.70 & 30.38 & 69.64 & 47.09 & 23.81 & 43.94 & 82.71\% \\
    GLA & 29.85 & 55.88 & 59.81 & 31.31 & 70.08 & 48.13 & 24.15 & 45.60 & 85.84\% \\
    RetNet & 26.06 & 54.06 & 56.86 & 30.29 & 68.99 & 46.23 & 23.23 & 43.67 & 82.21\% \\
    DeltaNet & 23.02 & 51.78 & 54.38 & 28.92 & 68.34 & 44.61 & 19.07 & 41.45 & 78.02\% \\
    Linear Attn & 16.95 & 50.99 & 49.33 & 25.60 & 65.51 & 37.74 & 14.37 & 37.21 & 70.05\% \\
    \midrule
    \rowcolor[gray]{.97} \multicolumn{9}{l}{\textbf{Scale: $\sim$1.7B Parameters}} \\
    SmolLM2-1.7B (Teacher) & 64.95 & 65.82 & 73.32 & 47.18 & 77.37 & 71.40 & 39.93 & 62.85 & 100.0\% \\
    Gated DeltaNet & 55.90 & 64.80 & 69.10 & 41.90 & 76.10 & 66.80 & 39.93 & 59.22 & 94.22\% \\
    xLSTM & 52.70 & 63.30 & 67.10 & 39.80 & 75.80 & 64.90 & 39.25 & 57.55 & 91.56\% \\
    Kimi & 42.20 & 57.50 & 60.70 & 35.30 & 73.40 & 59.70 &31.89 & 51.53 & 81.98\% \\
    GLA & 50.10 & 57.20 & 64.60 & 37.20 & 74.60 & 62.70 & 34.29 & 54.38 & 86.53\% \\
    RetNet & 50.00 & 58.60 & 65.80 & 38.80 & 74.80 & 62.90 & 32.44 & 54.76 & 87.13\% \\
    DeltaNet & 51.00 & 61.20 & 67.70 & 39.80 & 73.70 & 63.20 & 33.11 & 55.67 & 88.58\% \\
    Linear Attn & 30.90 & 53.70 & 55.10 & 31.30 & 69.60 & 52.50 & 18.94 & 44.58 & 70.92\% \\
    \bottomrule
\end{tabular}}
\caption{Experiment 1: Complete Zero-shot downstream benchmarks across all scales.}
\label{tab:appendix_all_results}
\end{table*}

\section{Full Results: Experiment 3}
\label{appendix:full_results}

This section provides detailed results for Experiment 3. We report results for the standard post-trained student models, aswell as the chat-distilled models.
For reference, we include results for a SmolLM2-1.7B model trained with the same regime (1 Epoch on Smoltalk) and the teacher model for the chat distillation.

\begin{table*}[h]
 \resizebox{\textwidth}{!}{
     \centering
     \begin{tabular}{lccccccr}
         \toprule
         \textsc{Model} & \makecell{\textsc{Lamb.} \\ {\small acc.}} & \makecell{\textsc{WinoG.} \\ {\small acc.}} & \makecell{\textsc{Arc-E} \\ {\small acc. norm.}} & \makecell{\textsc{Arc-C} \\ {\small acc. norm.}} & \makecell{\textsc{PIQA} \\ {\small acc. norm.}} & \makecell{\textsc{HellaS.} \\ {\small acc. norm.}} & \makecell{\textsc{BEAR} \\ {\small acc.}} \\ 
         \midrule
         \textbf{Post Training} \\
         Teacher (SmolLM2-1.7B, fine-tuned) & 61.61 & 67.25 & 66.46 & 42.58 & 76.55 & 70.82 & 36.94 \\
         Linear Attn & 26.11 & 52.09 & 45.16 & 29.35 & 66.81 & 48.67 & 25.21 \\
         Retnet & 48.01 & 61.56 & 55.85 & 36.01 & 74.05 & 62.26 & 32.19 \\
         GLA & 49.48 & 60.30 & 56.19 & 35.75 & 74.37 & 63.47 & 32.67 \\
         xLSTM & \textbf{54.62} & \textbf{65.43} & 63.59 & 37.63 & \textbf{75.30} & 66.58 & 37.49 \\
         Deltanet & 47.33 & 58.09 & 59.89 & 36.95 & 73.23 & 62.25 & 29.57 \\
         Gated Deltanet & 54.41 & 65.11 & \textbf{64.77} & \textbf{39.85} & \textbf{75.30} & \textbf{66.97} & 36.87 \\
         Kimi & 40.03 & 56.99 & 44.95 & 32.08 & 70.62 & 59.39 & 28.66 \\
         \midrule
         \textbf{Chat Distilled} \\
         \textit{Teacher (SmolLM2-1.7B-Instruct)} & 61.10 & 68.19 & 63.13 & 44.03 & 76.12 & 71.82 & 39.06 \\
         Linear Attn & 0.38 & 50.12 & 33.08 & 20.39 & 55.55 & 28.16 & 5.49 \\
         Retnet & 43.14 & 59.98 & 60.19 & 34.04 & 72.63 & 52.16 & 25.97 \\
         GLA & 28.28 & 54.22 & 55.85 & 32.42 & 69.15 & 54.75 & 17.60 \\
         xLSTM & 41.19 & 60.14 & 52.10 & 31.14 & 68.17 & 57.07 & 20.51 \\
         Deltanet & 41.89 & 58.41 & 59.97 & 37.80 & 72.42 & 61.70 & 27.21 \\
         Gated Deltanet & 41.93 & 57.93 & 60.10 & 37.97 & 72.36 & 61.83 & 27.04 \\
         Kimi & 12.48 & 50.59 & 53.32 & 32.42 & 68.06 & 49.43 & 11.29 \\
         \bottomrule
    \end{tabular}}
    \caption{Experiment 3: Performance of linearized models after post-training instruction tuning and instruction-aware distillation. Results are reported across instruction-following, reasoning, and long-context benchmarks.}
\end{table*}

\begin{table*}[h]
    \centering
    \resizebox{\textwidth}{!}{
    \begin{tabular}{lcccc|cccc|cccc}
    \toprule
    & \multicolumn{4}{c}{\makecell{\textsc{S-NIAH-1} \\ (pass-key retrieval)}} & \multicolumn{4}{c}{\makecell{\textsc{S-NIAH-2} \\ (number in haystack)}} & \multicolumn{4}{c}{\makecell{\textsc{S-NIAH-3} \\ (uuid in haystack)}}  \\
    Model & 512 & 1k & 2k & 4k & 512 & 1k & 2k & 4k & 512 & 1k & 2k & 4k \\
    \midrule
    \textbf{Post Training} \\
    Teacher (SmolLM2-1.7B, fine-tuned) & 100 & 100 & 100 & 100 & 100 & 100 & 99.8 & 99.8 & 100 & 99.8 & 97.4 & 95.0 \\
    Linear Attn & 0.0 & 0.0 & 0.0 & 0.0 & 0.0 & 0.0 & 0.0 & 0.0 & 0.0 & 0.0 & 0.0 & 0.0 \\
    Retnet & 12.4 & 1.0 & 0.0 & 0.0 & 47.8 & 13.8 & 0.4 & 0.0 & 0.0 & 0.0 & 0.0 & 0.0\\
    xLSTM   & 99.2 & 93.8 & 64.6 & 17.6 & 99.4 & 89.4 & 48.8 & 15.2 & 30.2 & 3.8 & 3.2 & 0.6 \\
    GLA   & 18.2 & 7.0 & 0.6 & 0.4 & 76.4 & 20.8 & 4.0 & 2.0 & 0.0 & 0.0 & 0.0 & 0.0 \\
    Deltanet & 98.6 & \textbf{99.4} & 0.0 & 0.0 & \textbf{99.6} & 92.8 & \textbf{62.4} & 0.0 & 18.2 & 1.6 & 0.0 & 0.0 \\
    Gated Deltanet & \textbf{99.8} & \textbf{99.4} & \textbf{98.0} & \textbf{65.8} & 99.4 & \textbf{95.0} & 58.2 & \textbf{21.6} & \textbf{32.2} & \textbf{12.0} & \textbf{4.6} & \textbf{2.0} \\
    Kimi & 57.2 & 9.6 & 2.6 & 0.6 & 97.4 & 25.4 & 11.4 & 4.2 & 0.0 & 0.0 & 0.0 & 0.0 \\
    \midrule
    \textbf{Chat Distilled} \\
    Teacher (SmolLM2-1.7B-Instruct) & 99.8 & 99.8 & 98.2 & 100 & 100 & 86.6 & 98.6 & 99.6 & 100 & 94.4 & 100 & 99.8 \\
    Retnet & 13.4 & 0.2 & 0.0 & 0.0 & 89.0 & 24.8 & 0.0 & 0.2 & 0.0 & 0.0 & 0.0 & 0.0 \\
    GLA & 1.4 & 0.0 & 0.0 & 0.0 & 49.4 & 1.4 & 0.0 & 0.0 & 0.0 & 0.0 & 0.0 & 0.0 \\
    xLSTM & 19.8 & 3.4 & 1.0 & 0.8 & 70.0 & 26.0 & 2.2 & 1.8 & 0.0 & 0.0 & 0.0 & 0.0 \\
    Deltanet & 98.0 & 94.2 & 63.2 & 22.2 & 99.0 & 90.8 & 50.6 & 11.6 & 0.6 & 0.0 & 0.0 & 0.0 \\
    Gated Deltanet & 98.6 & 98.0 & 72.4 & 24.0 & 99.6 & 91.8 & 59.0 & 17.0 & 7.6 & 0.0 & 0.0 & 0.0 \\
    Kimi & 0.4 & 0.0 & 0.0 & 0.0 & 52.2 & 1.2 & 0.0 & 0.0 & 0.0 & 0.0 & 0.0 & 0.0 \\
    \bottomrule
    \end{tabular}}
    \caption{Experiment 3: Full results for the NIAH-S task of the RULER benchmark.}
    \label{tab:my_label}
\end{table*}

\section{Efficiency Comparison: Prefilling and Decoding Speed}\label{appendix:bench}
\begin{figure*}[h]
    \centering
    \begin{subfigure}[t]{0.5\textwidth}
        \centering
        \includegraphics[width=0.99\textwidth]{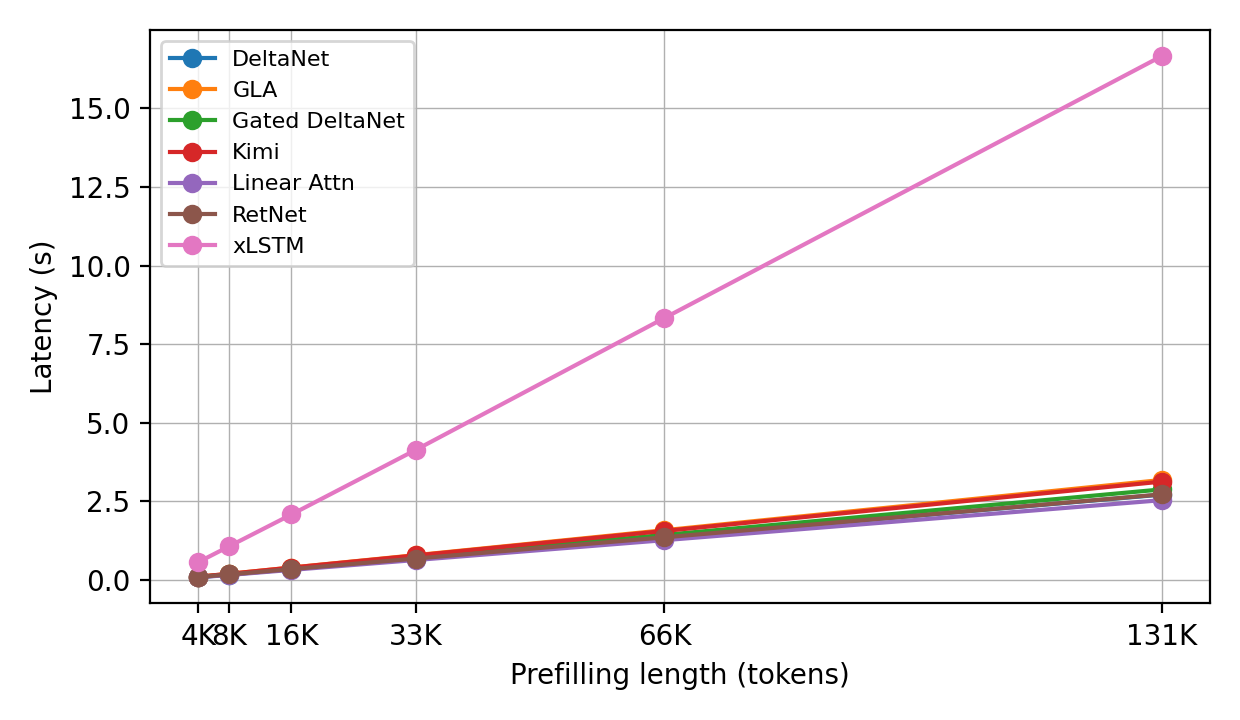}
        \label{fig:prefill}
    \end{subfigure}%
    ~ 
    \begin{subfigure}[t]{0.5\textwidth}
        \centering
        \includegraphics[width=0.99\textwidth]{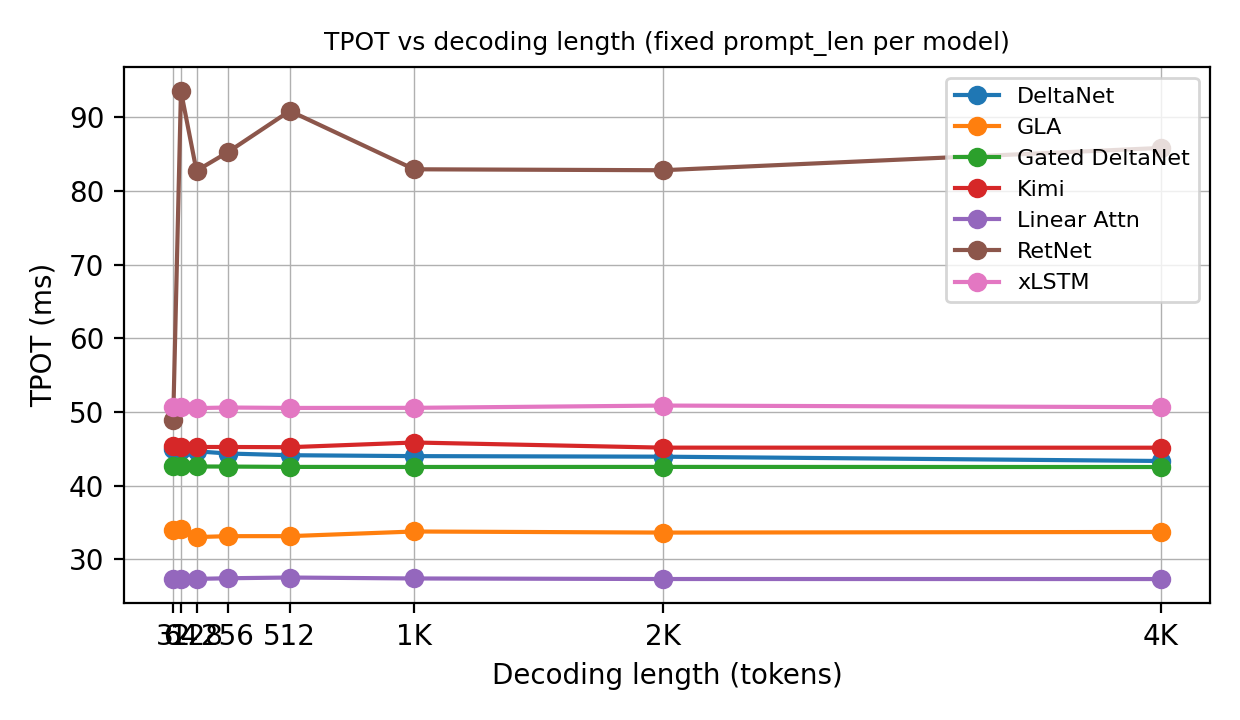}
        \label{fig:decode_time}
    \end{subfigure}
    \caption{The prefilling and decoding time of all 1.7B distilled models for different input and decoding lengths. All models exhibit linear scaling for ingesting input and constant time for generating tokens.}
    \label{fig:benchmark}
\end{figure*}

While each subquadratic architecture should scaling linearly with sequence length during decoding, we validate trainind and decoding times for all models and compare them against a standard transformer architecture (teacher model) in Figure~\ref{fig:benchmark}. All models are based on the SmollM2-1.7B teacher model with comparable parameter count.

\newpage

\end{document}